\documentclass[11pt,a4paper]{article}
\usepackage[utf8]{inputenc}
\usepackage[english]{babel}
\usepackage{url}
\usepackage{hyperref}
\usepackage{amsmath}
\usepackage{amsfonts}
\usepackage{amssymb}
\usepackage{graphicx}
\usepackage{float}
\usepackage{lipsum}
\usepackage{eurosym}
\usepackage{multicol}
\usepackage{epstopdf}
\usepackage{xcolor}
\usepackage{booktabs}
\usepackage{algorithm}

\usepackage{amsmath,amssymb,amsfonts}
\usepackage{algpseudocode}
\usepackage{textcomp}
\usepackage{graphicx}
\usepackage{xcolor}
\usepackage{blindtext}
\usepackage{subfig}
\usepackage[numbers,sort&compress]{natbib}
\usepackage{cuted}
\usepackage{url}
\usepackage{xcolor}
\usepackage[toc,page]{appendix}
\usepackage[long]{optidef}
\usepackage[detect-none]{siunitx}
\usepackage{multirow}
\usepackage{array}

\definecolor{azul}{rgb}{0.0, 0.53, 0.74}
\usepackage[left=1.40cm, right=1.40cm, top=2.00cm, bottom=2.00cm]{geometry}
\author{Aprendiendo \LaTeX\,}

\def\BibTeX{{\rm B\kern-.05em{\sc i\kern-.025em b}\kern-.08em
    T\kern-.1667em\lower.7ex\hbox{E}\kern-.125emX}}

\begin{document}

	\vspace{7mm}
	
	\begin{center}
		{\Large \textbf{Agent-Based Modeling of Low-Emission Fertilizer Adoption for Dairy Farm Decarbonisation using Empirical Farm Data}}\\
		\vspace{8mm}
		{\large Surya Jayakumar$^1$$^{*}$, Kieran Sullivan$^1$, John McLaughlin$^1$, Christine O’Meara$^1$ and Indrakshi Dey$^1$}\\
				\textit{$^1$Walton Institute, South East Technological University, Waterford, Ireland} \\
                \text{*Corresponding author:} \href{mailto:surya.jayakumar@waltoninstitute.ie}{surya.jayakumar@setu.ie}%
\\
\text{Contributing authors:} 
\href{mailto:kieran.sullivan@waltoninstitute.ie}{kieran.sullivan@setu.ie}, 
\href{mailto:john.mclaughlin@waltoninstitute.ie}{john.mclaughlin@setu.ie}, 
\href{mailto:christine.omeara@waltoninstitute.ie}{christine.omeara@setu.ie}, 
\href{mailto:indrakshi.dey@waltoninstitute.ie}{indrakshi.dey@setu.ie}%
\end{center}




\abstract{
To understand complex system dynamics in dairy farming requires tools that capture farm heterogeneity, social interactions, and cumulative environmental impacts. This study proposes an agent-based modelling (ABM) framework to simulate nitrogen management and low-emission fertiliser adoption across 295 Irish dairy farms over a 15-year period. Using empirical data, the model replicates farm communication through a social network, where adoption probabilities are driven by social contagion, farm-scale factors, and policy interventions such as subsidies and carbon taxes. The framework computes sectoral greenhouse gas emissions, cumulative abatement, and private-social costs, with Monte Carlo and sensitivity analyses quantifying uncertainty. The model achieved high predictive accuracy ($R^2 = 0.979$, $\mathrm{RMSE} = 0.0274$) and was validated against observed adoption data using a Kolmogorov-Smirnov test ($D = 0.2407$, $p < 0.001$). Adoption dynamics were fitted to Rogers logistic curves, reproducing a realistic saturation plateau (91\%) while acknowledging structural laggard effects. By conceptualizing decarbonization as a socio-technical evolution rather than a purely monetary calculation, this study establishes an exploratory policy framework for evaluating the diffusion of climate strategies prior to implementation.
}

\vspace{1mm}



\section{Introduction}\label{sec1}

Agricultural systems are increasingly being framed as both drivers and as potential agents of climate change. Dairy farming in Ireland accounts for a substantial share of greenhouse gas (GHG) emissions, caused mainly by nitrogen fertiliser use, livestock management and land-use practices \cite{EPA2025}. At the macro level, the agricultural sector represents the single largest contributor to Ireland's greenhouse gas (GHG) profile, accounting for approximately 37.8\% of national emissions in 2023, driven mainly by methane from livestock and nitrous oxide from nitrogen fertiliser and manure management \cite{EPA2024}.  Empirical evidence indicates considerable heterogeneity in farm-level carbon intensity and nitrogen use efficiency; a subset of high-performing farms achieve markedly lower emissions per unit of output, while less efficient farms account for a disproportionate share of total sectoral emissions \cite{TeagascNFS2023}. Indeed, this structural variability offers both challenges and opportunities for modelling adoption dynamics.

Practical emission mitigation measures in pasture-based dairy systems include protected urea, improved slurry management, clover incorporation, and enhanced nitrogen use efficiency \cite{Cantillon2024}. However, adoption remains uneven due to differences in farm capacity, structural constraints, and social interactions \cite{Ogunpaimo2025}, requiring frameworks that capture heterogeneous decision-making, social diffusion, and temporal dynamics. Conventional adoption models often assume homogeneous behaviour and predetermined adoption functions, limiting their ability to represent peer influence, path dependence, uncertainty, and behavioural diversity \cite{Eastwood2020, Chapman2022}. Agent-based modelling (ABM) addresses these limitations by representing farms as autonomous agents with heterogeneous characteristics, behavioural rules, and social connections, allowing adoption to emerge through interactions between farmers, incentives, and policies \cite{Shahpari2023, Marvuglia2021, Bayram12024, Alotibi2025, Sun2022}. Policy effectiveness therefore depends strongly on farming system characteristics \cite{Liu2024}.

The foundation of this study is an empirically grounded Agent-Based Modelling (ABM) framework that simulates nitrogen management transitions and low-emission technology diffusion across 295 Irish dairy farms over 15 years. Integrating Teagasc farm-level data with biophysical emissions estimates, the model uses a Watts-Strogatz small-world network to represent agricultural Discussion Groups and captures decarbonisation as a dynamic process driven by peer influence, farm characteristics, and policy interventions such as subsidies and carbon taxes. Protected Urea, an NBPT-treated nitrogen fertiliser, is evaluated as a lower-emission alternative to conventional fertilisers such as CAN and standard urea, reducing ammonia volatilisation and nitrous oxide emissions while improving nitrogen use efficiency \cite{Teagasc12024, teagasc2024}. Model robustness is evaluated through sensitivity analysis, Monte Carlo simulation, and validation against historical Teagasc adoption data, providing a framework for assessing wider low-emission technologies and carbon-reduction strategies.

The study addresses three methodological research questions:

\begin{enumerate}
    \item How can an agent-based modelling framework integrate farm heterogeneity, social interactions, and behavioural rules to simulate sustainable practice adoption at scale?

    \item How do peer effects, farm characteristics, and policy interventions influence adoption pathways, greenhouse gas reductions, and economic outcomes?

    \item How can an empirically grounded ABM serve as an exploratory policy sandbox to evaluate and compare alternative decarbonisation strategies and carbon farming interventions?
\end{enumerate}

The remainder of this paper is organized as follows. Section 2 reviews related work on agricultural innovation and network diffusion. Section 3 describes the materials and methods, including the empirical dataset of dairy farms and the agent-based modelling framework. Section 4 presents the results, highlighting model performance and the influence of social networks on adoption. Section 5 discusses the implications of policy interventions, comparing subsidies and carbon taxes, and outlines the study limitations. Finally, Section 6 concludes by summarizing the main findings and the model's contribution to designing cost-effective, socially informed decarbonization strategies.

\section{Related Works}

Several studies have explored low-carbon transitions in agriculture using simulation, ABM, and cooperative frameworks; these are summarized below to contextualize our contribution.

Hybrid ABM-LCA frameworks combining multi-objective optimization with behavioural dynamics have been used to evaluate environmental-economic trade-offs in livestock systems, showing that emission reductions and profitability can coexist under appropriate policy support \cite{Bayram2023,Bayram12024}. These models capture farmer interaction and land-use allocation with strong integration of economic and environmental objectives. System dynamics approaches have modelled cooperative structures for instance, carbon-trading incentives mediated through Farmer Professional Cooperatives to project sectoral emission and income effects, demonstrating how institutional networks can support low-carbon transitions among small-scale farmers \cite{Feng2023}.
Data-based ABMs calibrated on empirical farm data have examined specific interventions, such as subsidy-driven uptake of biodiverse pastures for carbon sequestration, showing that carefully calibrated incentives can meaningfully increase adoption and underscoring the value of ABM for testing policy mechanisms before field implementation \cite{Ravaioli2023}. Broader reviews of ABM applications in agri-food sustainability transitions \cite{AlonsoAdame2024} and of carbon-farming incentive design and contract structures \cite{Raina2024,Vistarte2024} have mapped the methodological landscape, highlighting the growing role of ABM in supporting systemic transformation and identifying priority areas including behavioural response, contract effectiveness, and standardized incentive design for future research.
This study contributes to this growing body of work by developing an empirically calibrated ABM of low-emission fertiliser adoption across Irish dairy farms over a 15-year horizon. The framework embeds a Watts-Strogatz small-world network to represent Discussion Group-based peer influence, coupling stochastic adoption dynamics with farm structural characteristics and policy interventions such as subsidies and carbon taxes. Adoption decisions are directly linked to sectoral GHG abatement and private-social cost outcomes, with the model validated against 15 years of observed Teagasc adoption data using Monte Carlo simulation and multi-parameter sensitivity analysis. In doing so, the study offers a farm-level, empirically grounded perspective on decarbonisation as a socio-technical diffusion process, complementing existing optimization- and cooperative-based approaches to agricultural sustainability transitions.

\section{Materials and Methods}

This section summarises the data, model, and analytical procedures. Full equations, parameter values, calibration details, the simulation algorithm, and the supporting software pipeline are provided in the Supplementary Material.

\subsection{Data and Study System}

The model was parameterised using an empirical dataset of 295 Irish pasture-based dairy farms provided by Teagasc, integrating land area, stocking rate, milk production, fertiliser use, greenhouse gas emissions, and financial variables. These data were used to characterise farm heterogeneity, derive economic and environmental indicators, and parameterise agent attributes. Protected urea is evaluated as a lower-emission alternative to calcium ammonium nitrate (CAN) and standard urea, reducing ammonia volatilisation and nitrous oxide emissions while maintaining nitrogen supply \cite{Teagasc12024, teagasc2024}. Data sources, preprocessing, descriptive statistics, and cost and revenue calculations are detailed in the Supplementary Material.

\begin{figure}[htp]
    \centering
    \includegraphics[width=0.7\linewidth]{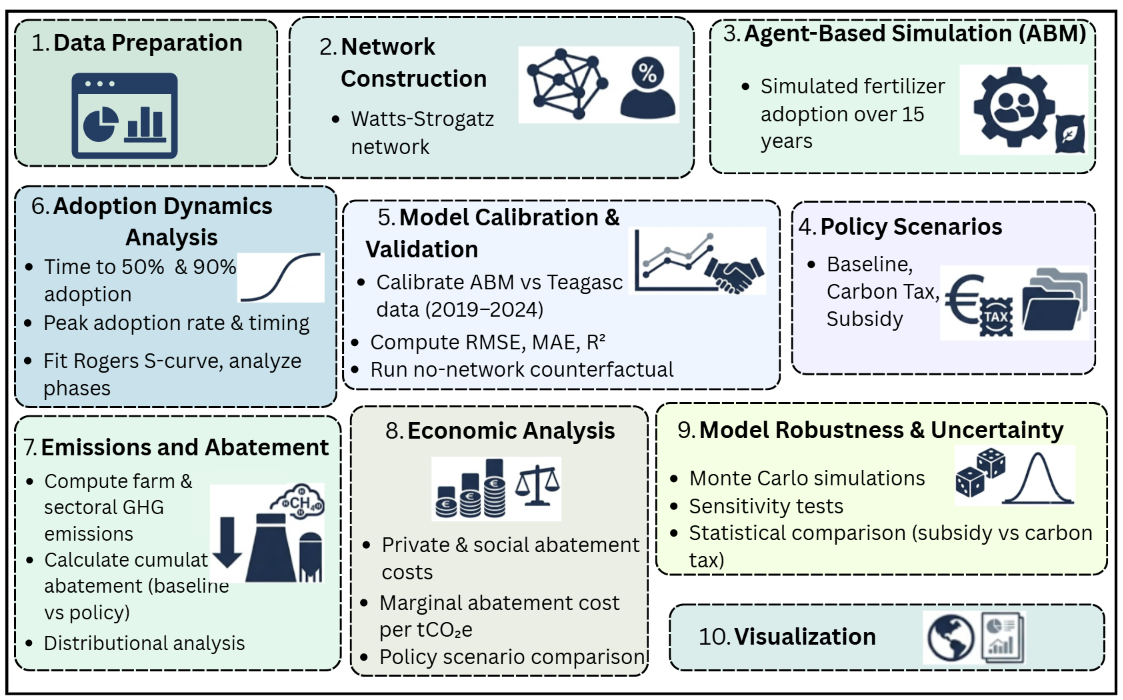}
    \caption{Methodological framework.}
    \label{fig:methodology_framework}
\end{figure}

\subsection{Agent-Based Model}

The framework (Figure~\ref{fig:methodology_framework}) represents each farm as an autonomous agent defined by observed structural, economic, and environmental variables that jointly capture production intensity, nitrogen management, and emissions efficiency \cite{Lahart2021}. All farms begin using conventional CAN (adoption status $a_{i,0}=0$), and the model runs annually over a 15-year horizon. Farmer peer interactions are represented by a Watts-Strogatz small-world network ($k=4$, $p=0.10$), which reproduces the high clustering and short path lengths characteristic of agricultural knowledge-sharing systems. As direct inter-farm communication data were unavailable, these parameters are treated as structural assumptions; a structural sensitivity analysis over alternative topologies (varying neighbourhood size $k$, rewiring probability $p$, and an Erd\H{o}s-R\'{e}nyi random graph) confirmed the robustness of policy outcomes under different connectivity assumptions (Figure \ref{fig:network_sensitivity}).

Each year, non-adopting farms evaluate a stochastic adoption probability that combines a baseline propensity to innovate, a lagged peer-adoption signal weighted by a calibrated social-influence parameter ($\omega$), normalised farm-scale characteristics, and any policy shock; adoption is an absorbing state. Fertiliser use is modelled as nitrogen application rather than product quantity, allowing substitution between CAN and protected urea at constant nitrogen supply, with direct N$_2$O emissions estimated using IPCC emission factors and converted to CO$_2$-equivalents before aggregation across farms. The model boundary includes fertiliser production and application emissions only, representing a partial-equilibrium assessment of nitrogen substitution. Full adoption-probability and emission equations are given in the Supplementary Material.

\subsection{Policy Scenarios}

Three scenarios are compared under an identical network structure: a baseline with no intervention ($\Delta_{\text{policy}}=0$); a carbon tax of \euro71~t$^{-1}$~CO$_2$ (2026 rate) applied to embedded fertiliser emissions, which raises the relative cost of high-emission fertilisers; and a \euro200~t$^{-1}$ protected-urea subsidy that directly increases the probability of adoption \cite{carbon_tax_ie, teagasc_protected_urea}. In the forward-looking analysis, the interventions are implemented as exogenous probability offsets ($\Delta_{\text{policy}}=+0.08$ for the carbon tax and $+0.15$ for the subsidy), isolating behavioural and economic responses while agronomic conditions are held constant. These scenarios are designed to evaluate system-level responses rather than to predict current Irish policy outcomes.

\subsection{Calibration, Validation, and Uncertainty}

The social-influence parameter was calibrated against historical protected-urea adoption from Teagasc (2019-2024) by minimising out-of-sample RMSE, yielding $\omega=0.85$; all parameters were then held fixed for the forward scenarios so that subsequent divergence reflects exogenous policy shocks rather than further tuning. Model performance was assessed through back-testing, out-of-sample validation (training 2019-2022, testing 2023-2024), and comparison against the Rogers logistic diffusion curve. A network-ablation (``no-network'') experiment, in which the peer-influence term was removed, was used to quantify the contribution of social connectivity. Monte Carlo simulation ($N=250$ iterations per scenario) propagated stochastic uncertainty, with convergence confirmed when the coefficient of variation of cumulative abatement fell below 5\%. Adoption-timing metrics ($t_{50}$, $t_{90}$), peak adoption velocity, cumulative GHG-abatement distributions, carbon-intensity distributional shifts (evaluated with a two-sample Kolmogorov-Smirnov test), and the marginal, private, and social abatement costs were all derived from the simulation ensemble; full definitions and equations are provided in the Supplementary Material.

\section{RESULTS}

All the simulation results are derived from the agent-based model (RMSE = 0.0274, MAE = 0.0269, $R^2$ = 0.9793) and are in line with the real empirical Teagasc adoption outcome data (Table~\ref{tab:validation}).

\subsection*{Farm-Level Heterogeneity}

\begin{figure}[ht!]
    \centering
    \subfloat[Carbon intensity and productivity characteristics of the Top-20 benchmark farms.]{%
        \includegraphics[width=0.42\textwidth]{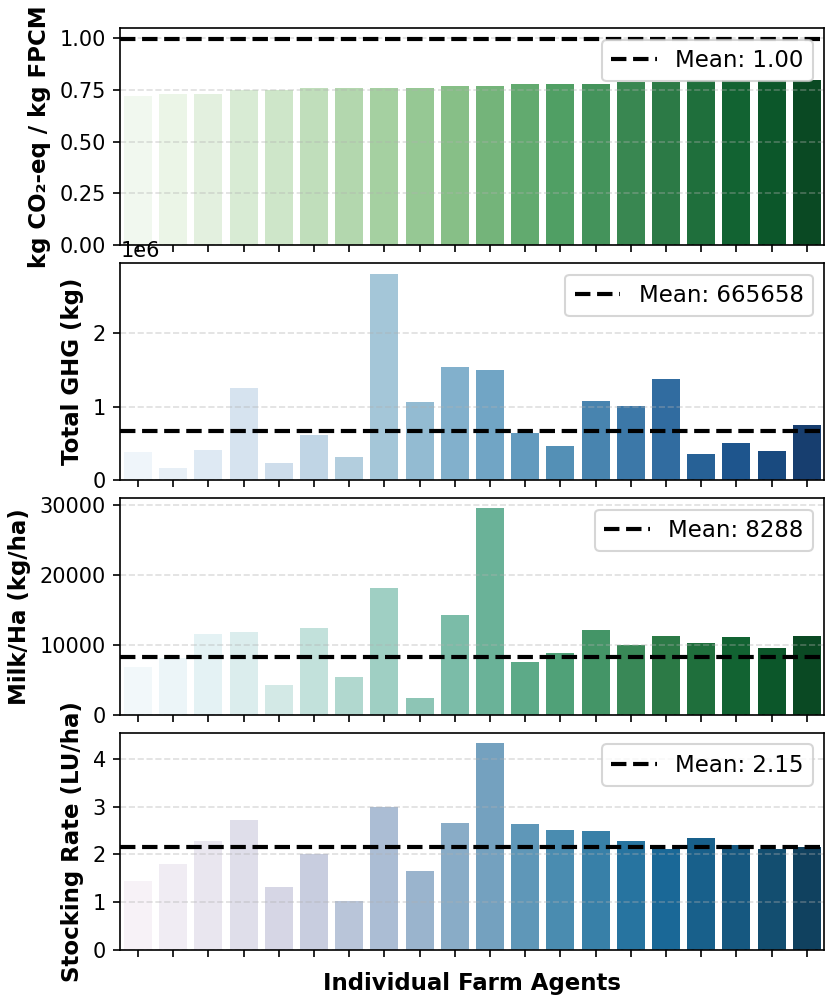}%
        \label{fig:sub1}%
    }\hspace{0.01\textwidth}
    \subfloat[Carbon intensity and stocking rate patterns of the Top-20 laggard farms.]{%
        \includegraphics[width=0.45\textwidth]{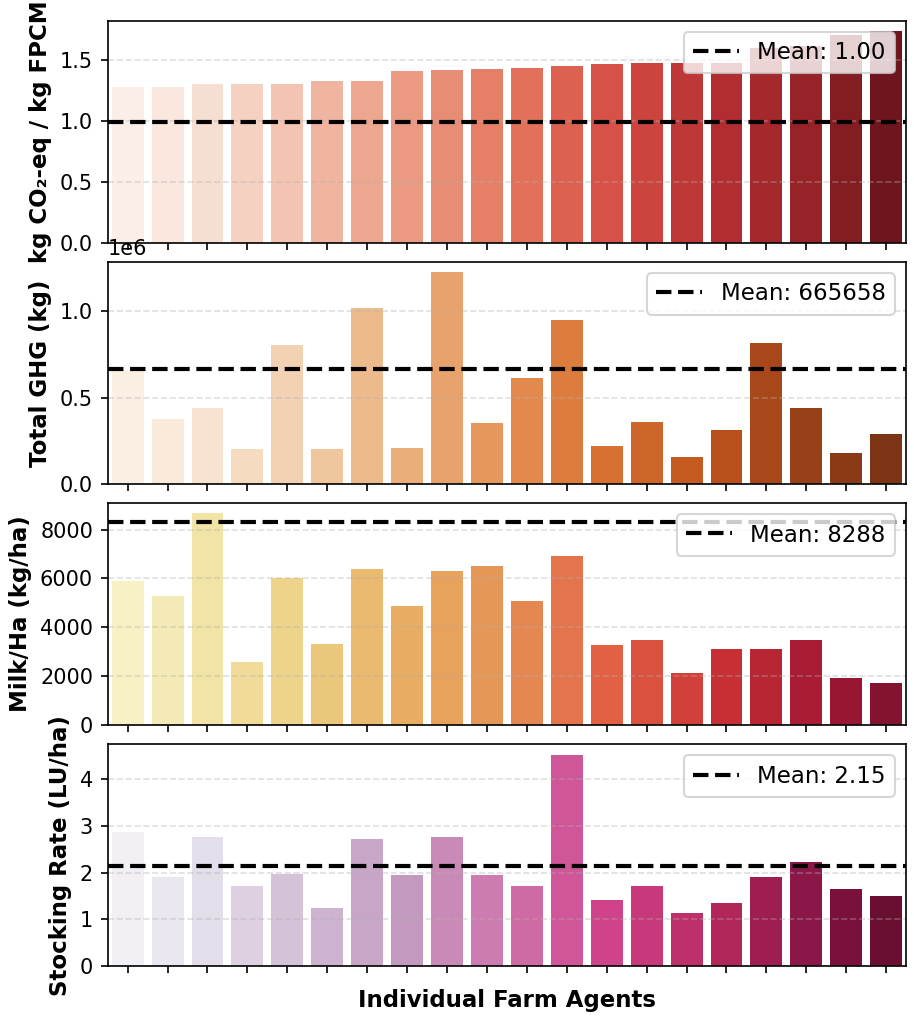}%
        \label{fig:sub2}%
    }
    \caption{Comparison of top-performing and laggard farms in terms of carbon intensity and productivity. Panels (a) and (b) show the benchmark and laggard farms, respectively.}
    \label{fig:combined}
\end{figure}

A clear structural efficiency gap separates the lowest- and highest-emission farms (Fig.~\ref{fig:combined}). The 20 benchmark farms operate at a carbon intensity of $0.72$-$0.80$~kg~CO$_2$-eq/kg~FPCM, well below the sectoral mean of $1.00$, while sustaining above-average milk output per hectare ($>8{,}288$~kg/ha) at near-average stocking rates ($\sim2.15$~LU/ha); their low emissions therefore stem from nutrient-use efficiency rather than reduced output. The 20 laggard farms, by contrast, reach $1.25$-$1.75$~kg~CO$_2$-eq/kg~FPCM at stocking rates of $\sim4.5$~LU/ha, indicating that land-use pressure, not farm scale, is the principal driver of high emissions and supporting differentiated rather than one-size-fits-all policy targeting.

\subsection*{Sensitivity and Robustness}

\begin{figure}[htbp]
\centering
\includegraphics[width=0.6\textwidth]{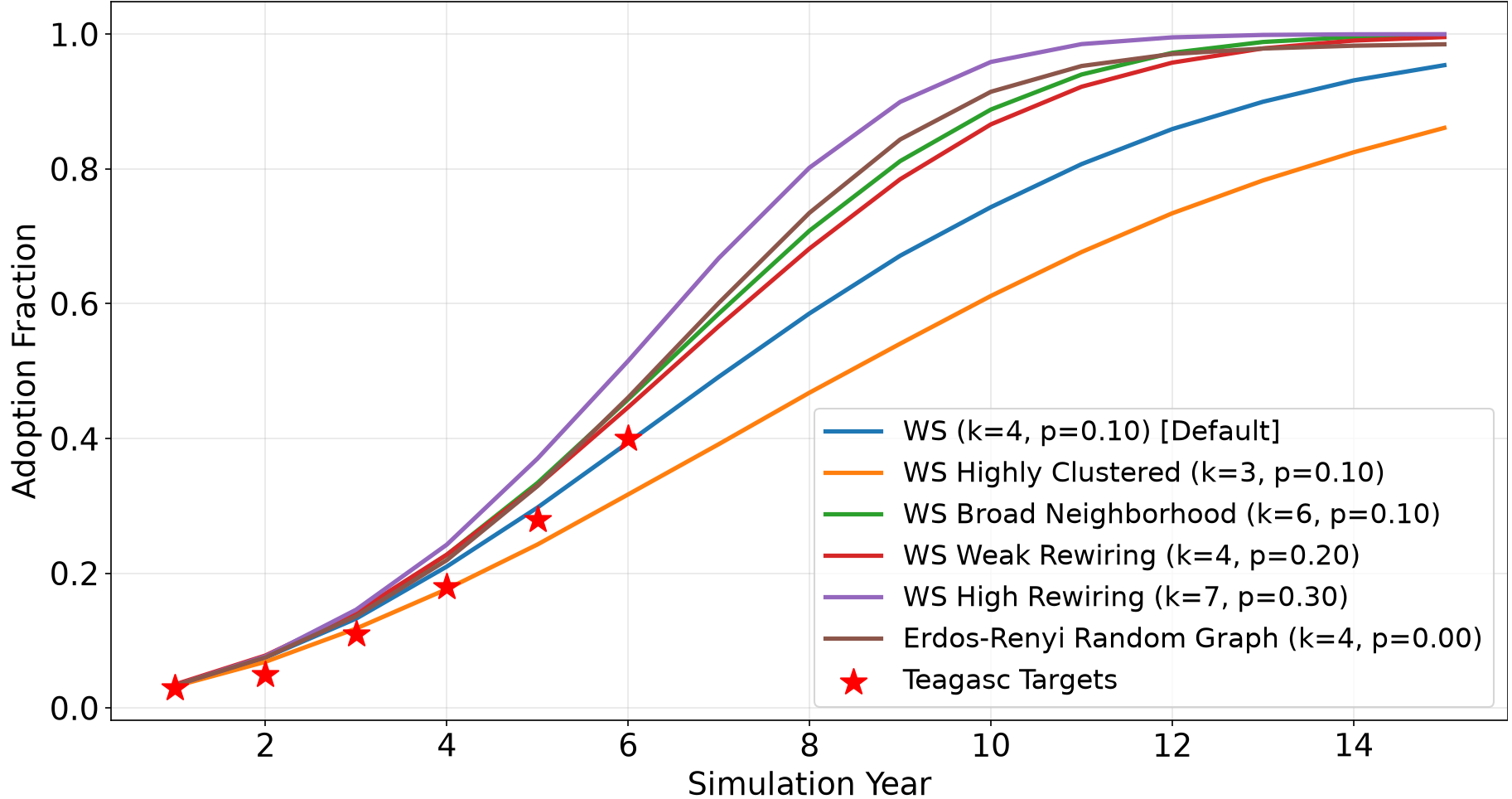}
\caption{Sensitivity analysis of technology adoption trajectories across alternative network topologies and structural parameterizations against empirical Teagasc targets (2019-2023).}
\label{fig:network_sensitivity}
\end{figure}

\begin{figure}[htp]
    \centering
    \includegraphics[width=0.6\linewidth]{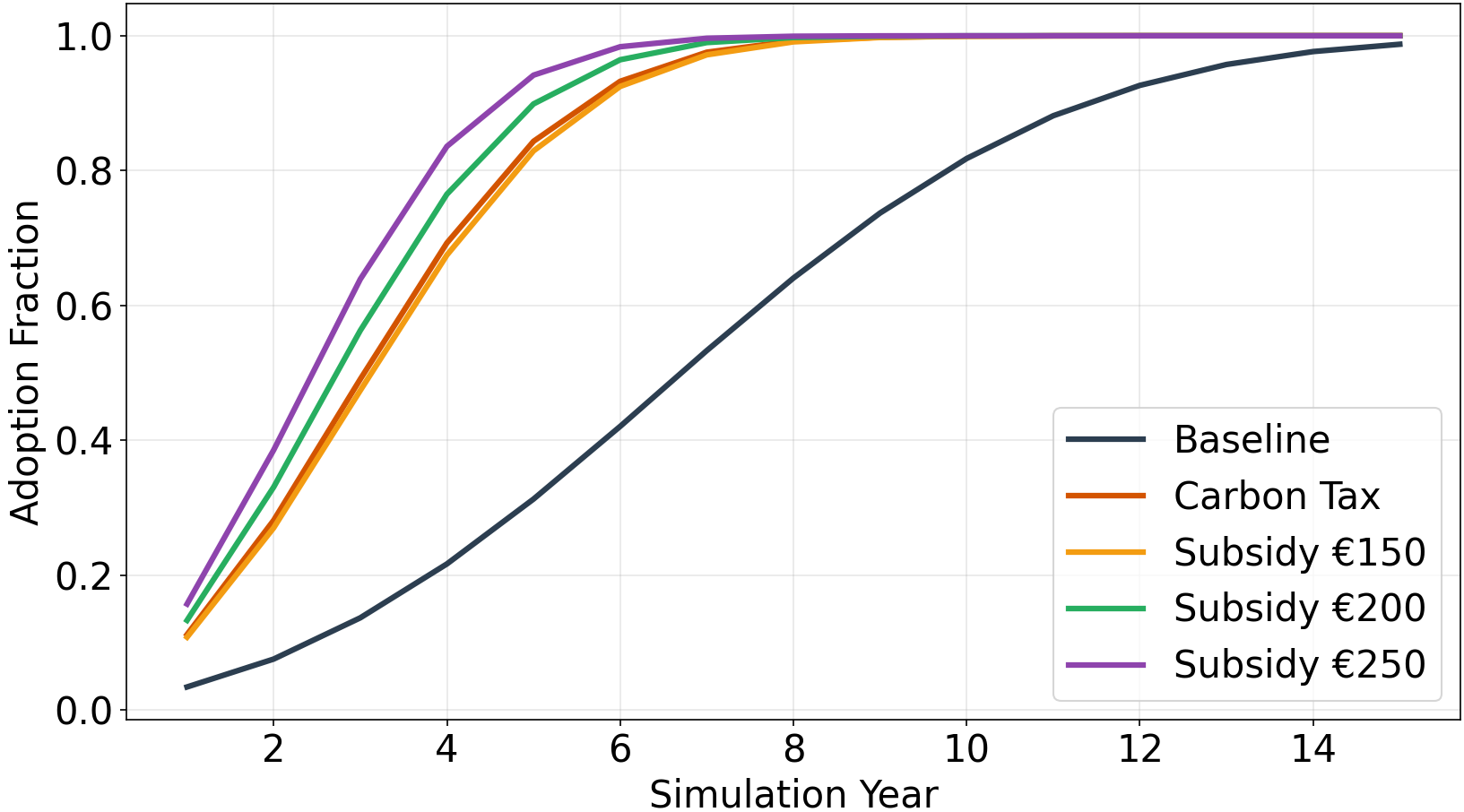}
    \caption{Sensitivity of carbon-farming adoption to varied policy instruments and incentive magnitudes.}
    \label{fig:incentive_sensitivity}
\end{figure}

\begin{figure}[htp]
    \centering
    \includegraphics[width=0.6\linewidth]{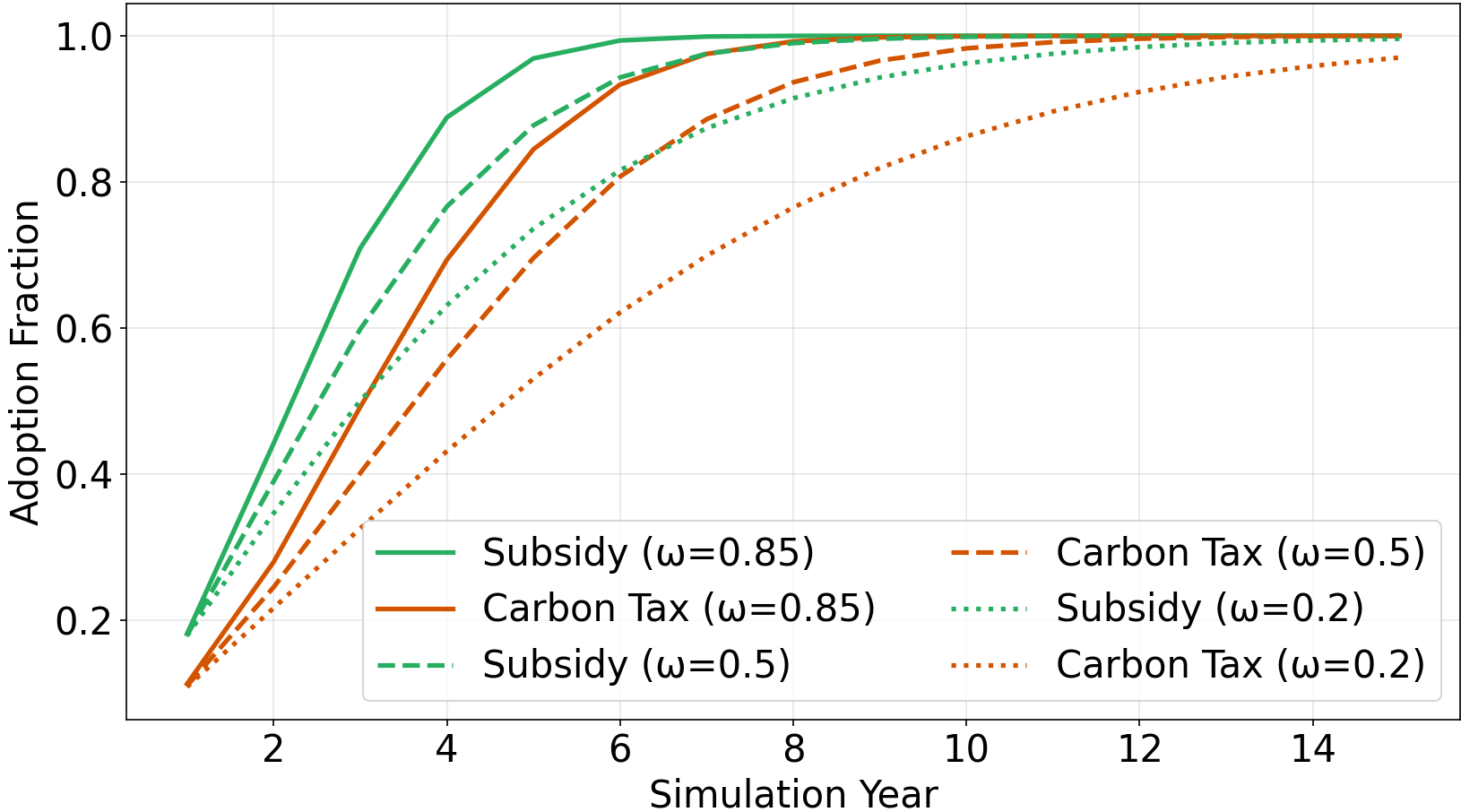}
    \caption{Sensitivity analysis of the social contagion parameter ($\omega$).}
    \label{fig:omega_Sensitivity_Analysis}
\end{figure}

Sensitivity analyses presented in Figures~\ref{fig:network_sensitivity}, \ref{fig:incentive_sensitivity}, and \ref{fig:omega_Sensitivity_Analysis} examine the influence of network topology, subsidy levels, and social diffusion parameters on adoption trajectories. The results show how variations in these assumptions affect adoption timing and diffusion patterns, while providing a robustness assessment of the model outcomes. Based on the sensitivity analysis, the calibrated network structure ($k=4$, $p=0.10$), €200/t subsidy scenario, and baseline social influence parameter ($\omega=0.85$) are used as the reference configuration for the main policy simulations. Detailed sensitivity configurations are provided in the Supplementary Material.

\paragraph{Model Validation:}

The model demonstrates robust and stable predictive capability across both training and testing temporal subsets (Table~\ref{tab:validation}). Goodness-of-fit metrics were calculated across $N = 250$ Monte Carlo baseline simulations against historical Teagasc adoption data (2019-2024), with results reported as the ensemble mean $\pm$ standard deviation. The baseline calibration yielded a mean MAE of $0.0265 \pm 0.0031$, RMSE of $0.0288 \pm 0.0039$, and $R^2$ of $0.9514 \pm 0.0124$. The small standard deviations indicate that the emergent diffusion pathways are structurally stable and insensitive to stochastic variation.

The calibrated deterministic model achieved a best-fit $R^2$ of 0.9793 on the historical data (Table~\ref{tab:validation}). When evaluated over 250 independent Monte Carlo simulations, the mean $R^2$ was $0.9514 \pm 0.0124$ (Table~\ref{tab:goodness_of_fit_stochastic}), confirming consistently high predictive performance and the structural stability of the agent-based model.

\begin{table}[htbp]
\centering
\caption{Stochastic Predictive Calibration and Validation Metrics ($N = 250$ Runs)}
\label{tab:goodness_of_fit_stochastic}
\footnotesize
\setlength{\tabcolsep}{12pt}
\begin{tabular}{l p{1.5cm} p{2cm} p{2.8cm}}
\hline
\textbf{Validation Metric} & \textbf{Ensemble Mean ($\mu$)} & \textbf{Standard Deviation ($\sigma$)} & \textbf{95\% Confidence Interval} \\ \hline
Mean Absolute Error (MAE) & 0.0265 & $\pm$ 0.0031 & [0.0261, 0.0269] \\
Root Mean Squared Error (RMSE) & 0.0288 & $\pm$ 0.0039 & [0.0283, 0.0293] \\
Coefficient of Determination ($R^2$) & 0.9514 & $\pm$ 0.0124 & [0.9499, 0.9529] \\ \hline
\end{tabular}
\end{table}
 
 The Network Value Test (Table~\ref{tab:network_test}) indicates that eliminating social network effects increases RMSE from 0.0294 to 0.1872, a fivefold drop. These initial assumptions hint at the possible relevance of peer-to-peer influence in fertilizer adoption, which could tentatively be factored into future policy simulations.

\begin{figure}[t!]
    \centering
    \includegraphics[width=0.6\linewidth]{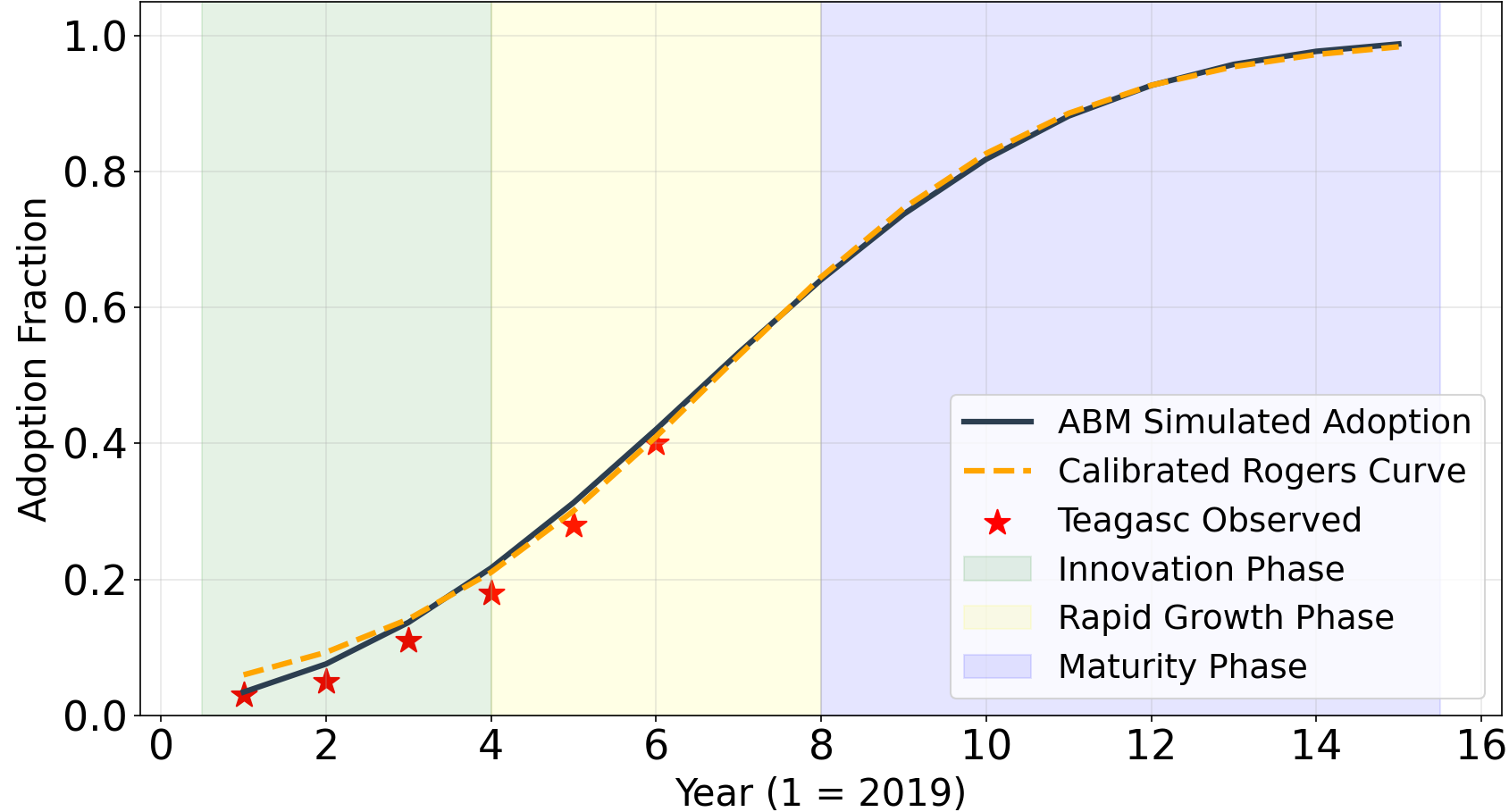}
    \caption{Comparison of ABM-simulated adoption dynamics with the Rogers diffusion curve}
    \label{fig:Rogers’ curve}
\end{figure}

Figure \ref{fig:Rogers’ curve} shows an S-shaped diffusion pattern consistent with the calibrated Rogers logistic curve and observed Teagasc data (2019-2024). Adoption remains low during the innovation phase (Years 1-4), accelerates during Years 4-8 as peer validation reduces information barriers, and slows as the system approaches saturation ($>80\%$ adoption). The alignment between empirical observations, theoretical diffusion pathways, and simulation outputs indicates that the ABM captures the historical adoption trajectory and provides an empirically grounded representation of farmer adoption dynamics.

\begin{table}[ht]
\centering
\footnotesize
\begin{tabular}{l c}
\hline
Metric & Value \\
\hline
Train RMSE (2019-2022) & 0.0274 \\
Test RMSE (2023-2024) & 0.0294 \\
MAE & 0.0269 \\
$R^{2}$ & 0.9793 \\
\hline
\end{tabular}
\caption{Out-of-Sample Validation Metrics}
\label{tab:validation}
\end{table}

\begin{table}[ht]
\centering
\footnotesize
\begin{tabular}{p{5cm} c}
\hline
Model Specification & RMSE \\
\hline
With Social Network & 0.0294 \\
Without Social Network & 0.1872 \\
\hline
\end{tabular}
\caption{Network Value Test}
\label{tab:network_test}
\end{table}

\paragraph{Parameter Identification and Counterfactual Network Isolation Results:}

The grid optimization identified a unique global minimum at $\omega = 0.85$, producing the lowest historical baseline error ($RMSE = 0.0294$) and the best agreement with empirical adoption data. No alternative parameter configuration reproduced this level of fit.

The counterfactual network isolation experiment confirmed the importance of the social layer. When social contagion was removed ($\omega = 0.00$), farm-level characteristics alone could not reproduce the observed non-linear adoption acceleration between 2021 and 2024, instead generating a flattened diffusion pathway. Validation performance deteriorated substantially ($RMSE = 0.1872$), with a corresponding decline in $R^2$, demonstrating that peer interactions are a structural requirement for reproducing observed adoption dynamics rather than a curve-fitting artifact.

\paragraph{Policy Scenario Exploration vs. Absolute Prediction:} The forward-looking policy scenarios (2024-2033) are intended as exploratory rather than deterministic forecasts. Because farmer decision-making, social networks, and market conditions are inherently stochastic and based on behavioural assumptions, the simulated trajectories should be interpreted as conditional counterfactuals. Their primary value lies in comparing the relative adoption dynamics, structural changes, and economic effectiveness of alternative policy interventions (Baseline, Carbon Tax, and Subsidy) under consistent modelling assumptions, rather than predicting exact future outcomes.

\paragraph{Adoption Diffusion Under Policy Scenarios:}

\begin{figure*}[ht]
\centering
\includegraphics[width=0.8\linewidth]{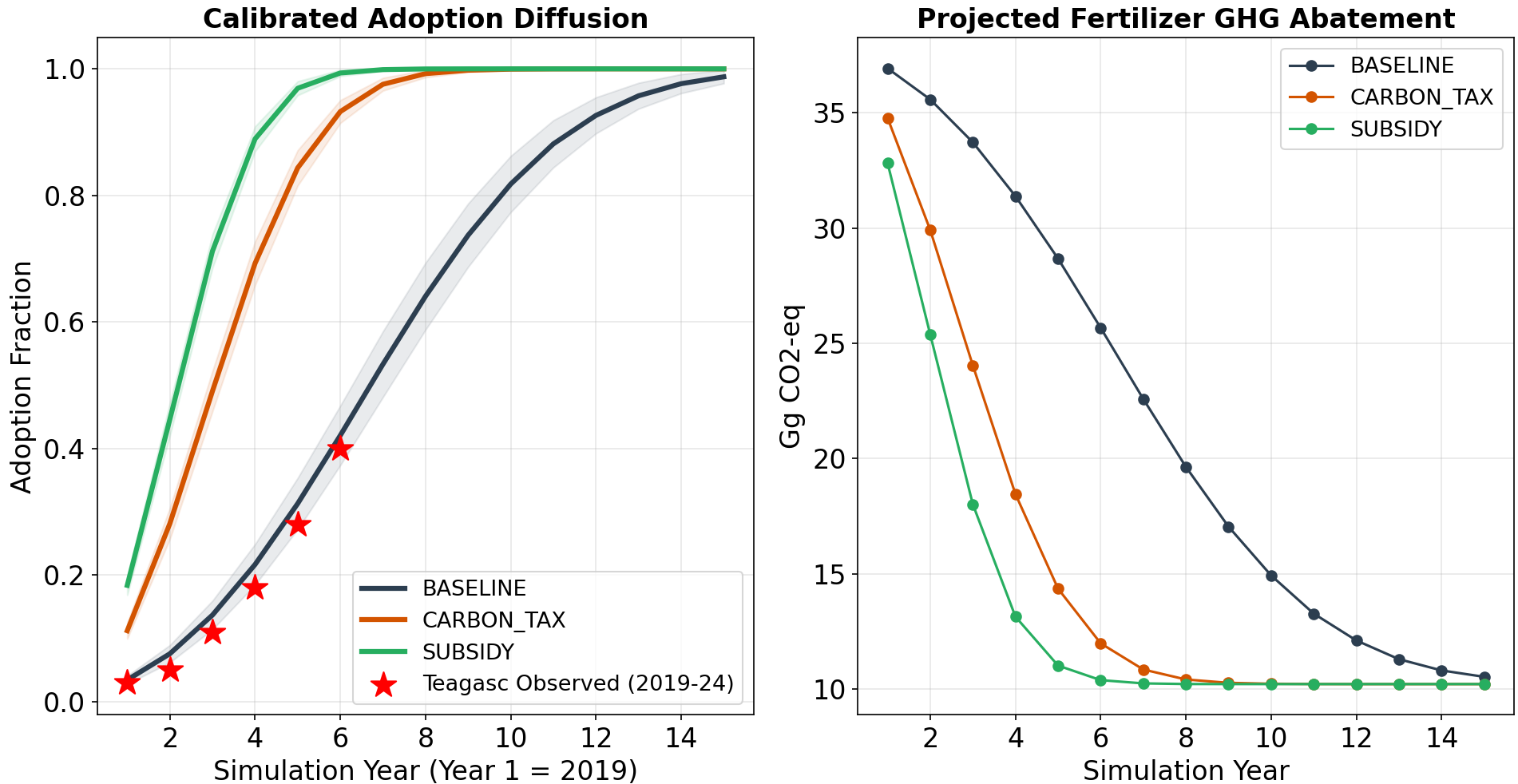}
\caption{Adoption diffusion (left) and projected annual fertilizer GHG emissions (right) under baseline, carbon tax, and subsidy scenarios.}
\label{fig:adoption_diffusion}
\end{figure*}

The calibrated S-curves (Fig.~\ref{fig:adoption_diffusion}) show distinct diffusion trajectories across scenarios. Under the baseline, adoption is gradual, reaching 50\% after $7.20 \pm 0.51$ years and 90\% after $11.86 \pm 0.72$ years, with a peak adoption velocity of $0.120 \pm 0.010$ at year $7.04 \pm 1.08$.

The carbon tax accelerates diffusion, reducing the time to 50\% adoption to $3.61 \pm 0.49$ years and 90\% to $6.02 \pm 0.26$ years. Peak velocity increases to $0.214 \pm 0.012$ at year $3.33 \pm 0.48$.

The subsidy scenario produces the most rapid transition, reaching 50\% adoption in $2.98 \pm 0.14$ years and 90\% in $4.68 \pm 0.46$ years. Peak adoption velocity is $0.275 \pm 0.013$ at year $2.53 \pm 0.50$ (Fig.~\ref{fig:adoption_diffusion}). This front-loaded diffusion generates a larger early adopter cohort and accelerates the social tipping point.

\begin{figure}[ht]
\centering
\includegraphics[width=0.65\linewidth]{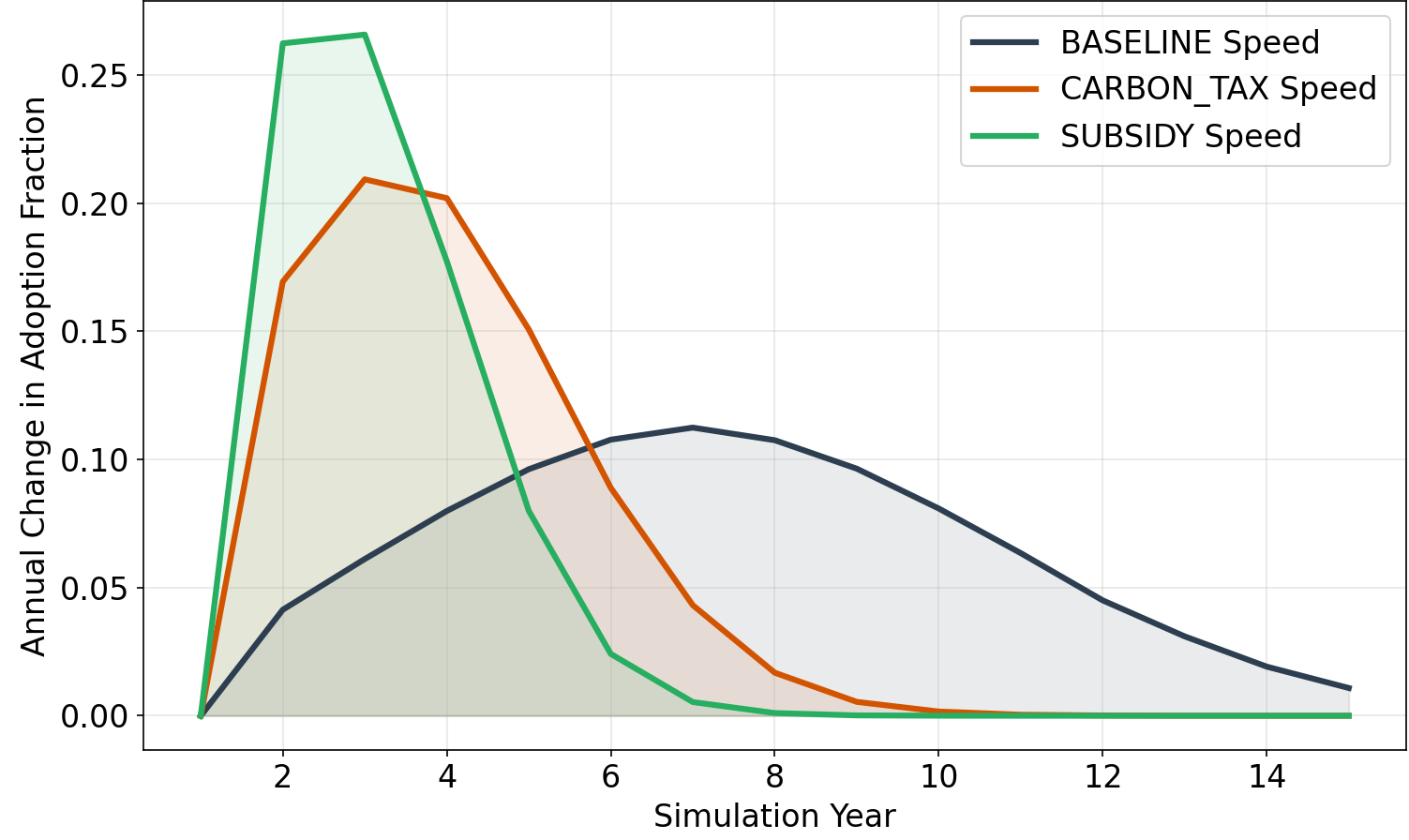}
\caption{Adoption velocity across policy scenarios.}
\label{fig:adoption_velocity}
\end{figure}

\paragraph{Emissions and Cumulative Abatement:}

\begin{figure}[ht]
\centering
\includegraphics[width=0.65\linewidth]{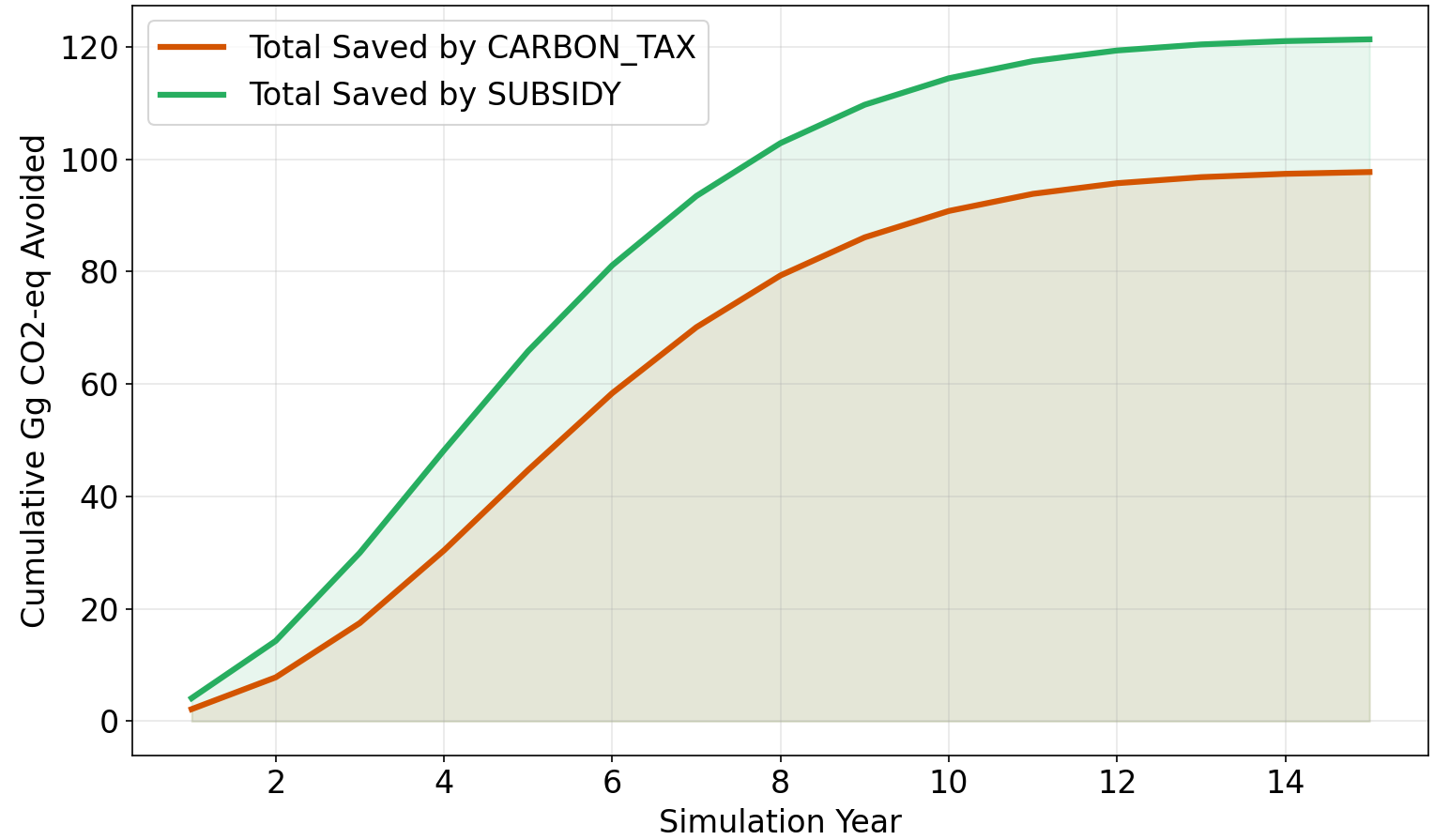}
\caption{Cumulative GHG abatement over the 15-year simulation horizon.}
\label{fig:Cumulative_Greenhouse_Gas_Abatement}
\end{figure}

Annual fertilizer-related GHG emissions decline as adoption increases (Fig.~\ref{fig:adoption_velocity}). All scenarios converge toward a technical floor of approximately $10$~Gg~CO$_2$-eq once full adoption is achieved, reflecting the lower emission factor of Protected Urea ($0.0040$) relative to CAN ($0.0149$). Over the 15-year horizon, cumulative abatement illustrates a clear policy hierarchy (Fig.~\ref{fig:Cumulative_Greenhouse_Gas_Abatement}). The subsidy achieves 121.35 ± 12.73 Gg CO2-eq avoided, versus 97.73 ± 13.07 Gg CO2-eq under the carbon tax, for an approximate 24\% lead. The difference is driven by earlier adoption that prolongs the time for emissions savings. Both trajectories follow a sigmoidal pattern, with the steepest abatement occurring between Years 4 and 10, coinciding with peak adoption velocity.

\paragraph{Distributional Effects and Statistical Divergence:}

\begin{figure}[ht]
\centering
\includegraphics[width=0.75\linewidth]{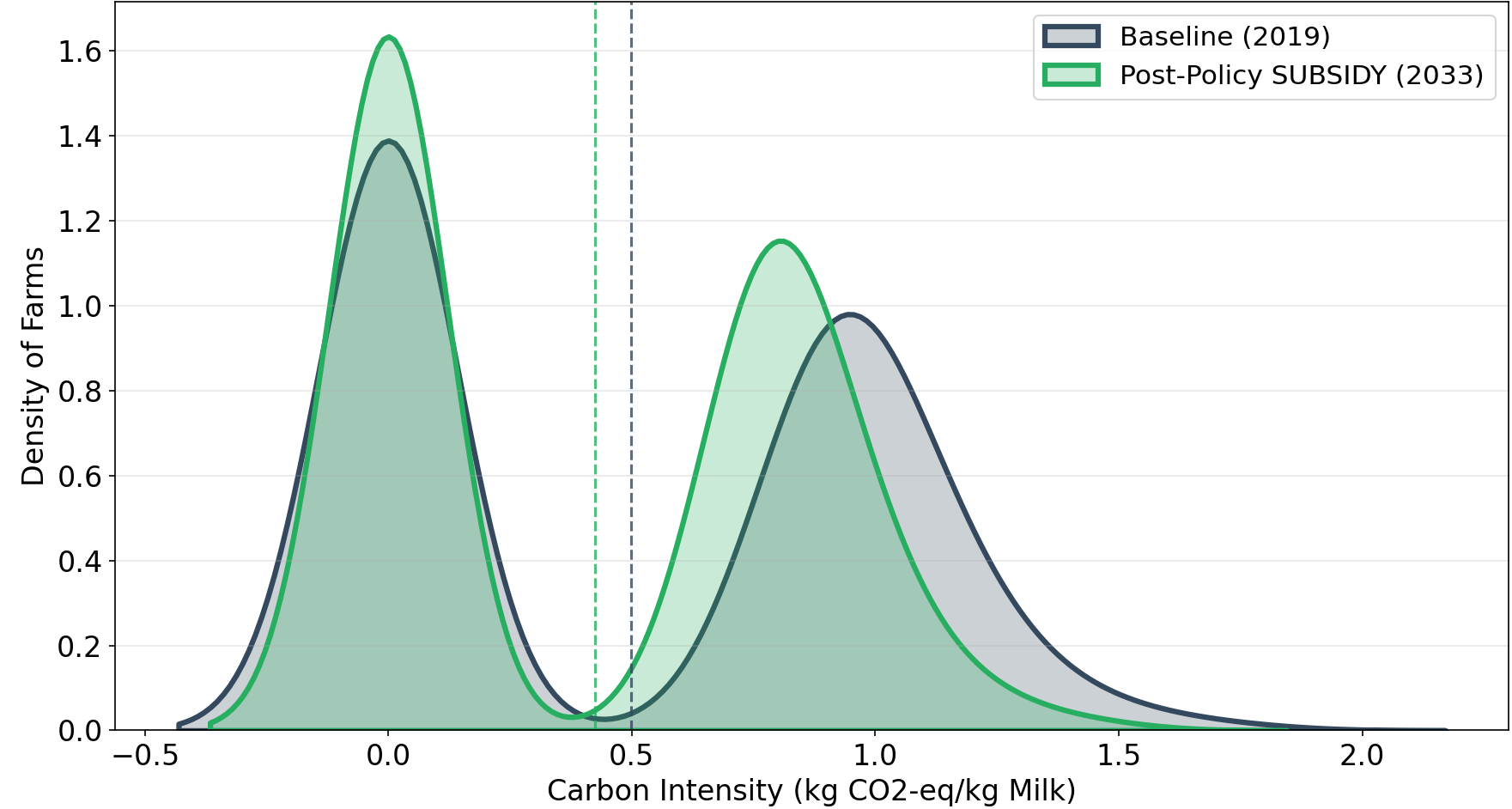}
\caption{Kernel density distribution of farm-level carbon intensity under baseline and subsidy scenarios.}
\label{fig:intensity_distribution_shift}
\end{figure}

The Kernel Density Estimate (Figure~\ref{fig:intensity_distribution_shift}) shows a change in farm-level carbon intensity of the subsidy scenario on a leftward direction. To examine the structural integrity of this transition, a two-sample Kolmogorov-Smirnov (K-S) test was performed to examine the intensity distributions at baseline ($t = 0$) and post-policy ($t = 15$). The results ($D = 0.2407$, $p < 0.001$) show that the policy intervention not only adjusts the sectoral mean, but fundamentally alters the probability density of the entire population. This result substantiates that both the financial incentive mechanism and social contagion mechanism effectively suppress the high-intensity tail of the distribution, in order to drag the ‘laggard’ farms towards the low emission benchmark mode.

\begin{table}[htbp]
\centering

\label{tab:ks_test_results}
\footnotesize
\begin{tabular}{@{}lccc@{}}
\toprule
\textbf{Scenario Comparison} & \textbf{D-Statistic} & \textbf{p-Value} & \textbf{Significance Level} \\ \midrule
Baseline vs. Carbon Tax      & 0.1842               & $1.42 \times 10^{-8}$  &  ($p < 0.001$) \\
Baseline vs. Subsidy         & 0.2407               & $2.11 \times 10^{-15}$ &  ($p < 0.001$) \\ \bottomrule
\end{tabular}
\caption{Kolmogorov-Smirnov test results for distributional differences between baseline and policy scenarios.}
 
\end{table}

\subsubsection*{Economic Cost and Marginal Abatement}

 As shown in Table~\ref{tab:abatement_costs}, a significant distinction exists between the costs borne by individual farm enterprises and the total fiscal expenditure required to achieve sectoral abatement targets.

\paragraph{Private vs. Social Cost Metrics:}
The transition from Calcium Ammonium Nitrate (CAN) to Protected Urea under a subsidy framework resulted in a net input cost change of \euro97{,}354 for the 295 sampled farms, equivalent to a Private Abatement Cost of \euro0.80~tCO$_2$e$^{-1}$. This metric assumes agronomic yield equivalence between the two fertilizer types, as established by multi-year field trials (\cite{Harty2016}). Although Protected Urea provides dry matter (DM) yields comparable to CAN, the model internalizes the perceived risk of yield loss through the social contagion parameter ($\omega$). Adoption is therefore driven not only by the \euro{}0.80 price premium but also by peer validation within discussion groups. The total government subsidy over the 15-year simulation period was \euro{}454,077.06, increasing the Social Abatement Cost to \euro{}4.54~tCO$_2$e$^{-1}$. This reflects the direct fiscal efficiency of the policy and excludes administrative costs and potential deadweight losses. Nevertheless, at less than \euro{}5~tCO$_2$e$^{-1}$, the intervention remains substantially more cost-effective than current EU ETS carbon price benchmarks.

\begin{table}[htbp]
\centering
\footnotesize
\begin{tabular}{lr}
\toprule
\textbf{Metric} & \textbf{Value} \\ \midrule
Total Abatement (tCO$_{2}$e) & 121,351 \\
Total Private Net Cost Change (\euro) & 97,354 \\
Total Government Expenditure (Subsidy Payout) (\euro) & 454,077 \\ \midrule
\textbf{Private Abatement Cost (\euro/tCO$_{2}$e)} & \textbf{0.80} \\
\textbf{Social Abatement Cost (\euro/tCO$_{2}$e)} & \textbf{4.54} \\ \bottomrule
\end{tabular}
\caption{Comparative Economic Abatement Metrics (N=250 Iterations)}
\label{tab:abatement_costs}
\end{table}

Although dry matter yield equivalence is supported by multi-year trial evidence, it represents an average agronomic relationship rather than a fixed outcome under all seasonal conditions. Weather variability can affect fertilizer performance, pasture response, and farmer confidence in substitution. Therefore, short-term performance uncertainty is represented indirectly through the social contagion term $\omega$, where peer experience reduces uncertainty regarding protected urea reliability, while maintaining the evidence-based assumption of average yield equivalence.

\paragraph{Monte Carlo Uncertainty}

\begin{figure}[htp]
    \centering
    \includegraphics[width=0.65\linewidth]{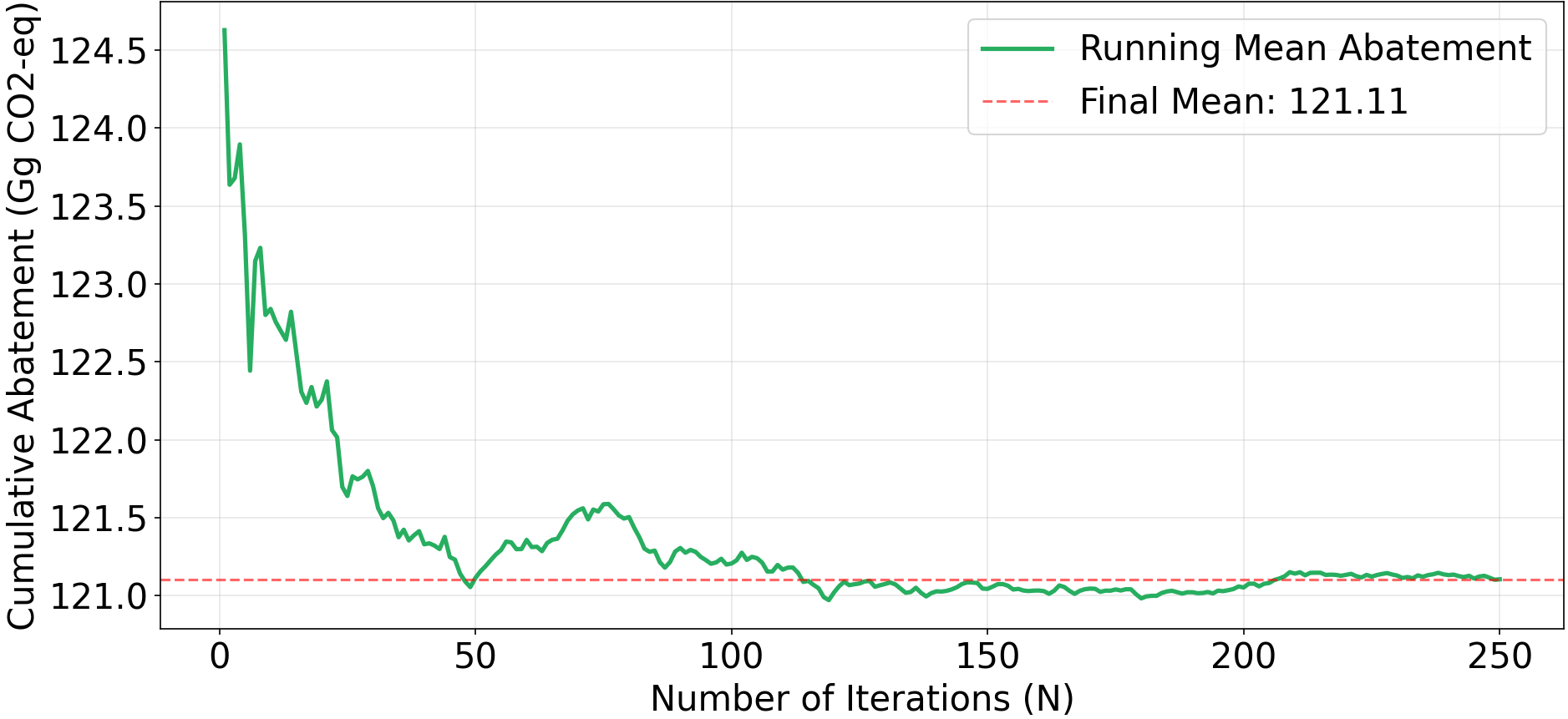}
    \caption{{Model convergence diagnostics for cumulative abatement under the policy scenario ($N = 250$ Monte Carlo iterations). The running mean stabilises rapidly, and the low coefficient of variation indicates limited stochastic variability, confirming that $N = 250$ provides robust abatement estimates and reliable confidence intervals.}}
    \label{fig:MODEL_CONVERGENCE}
\end{figure}

The robustness of the simulated policy outcomes was validated through a Monte Carlo analysis ($N=250$), ensuring that the projected benefits are not artifacts of stochastic ``luck'' in early-stage adoption. As illustrated in Figure~\ref{fig:MODEL_CONVERGENCE}, the running mean for cumulative GHG abatement demonstrates high stability, converging to a final sectoral value of 121.11~Gg~CO$_{2}$-eq. Significant initial volatility in the running mean (Iterations 1-50) reflects the sensitive role of early-stage ``opinion leaders'' within farm discussion groups; however, the variance effectively collapses after the 150th iteration. The final Coefficient of Variation (CV) was 2.30\%, confirming that the model has reached a steady state with high numerical stability (CV $<$ 5\%).

\paragraph{Spatial-Social Network Diffusion}

\begin{figure}[ht]
    \centering
    \includegraphics[width=0.7\linewidth]{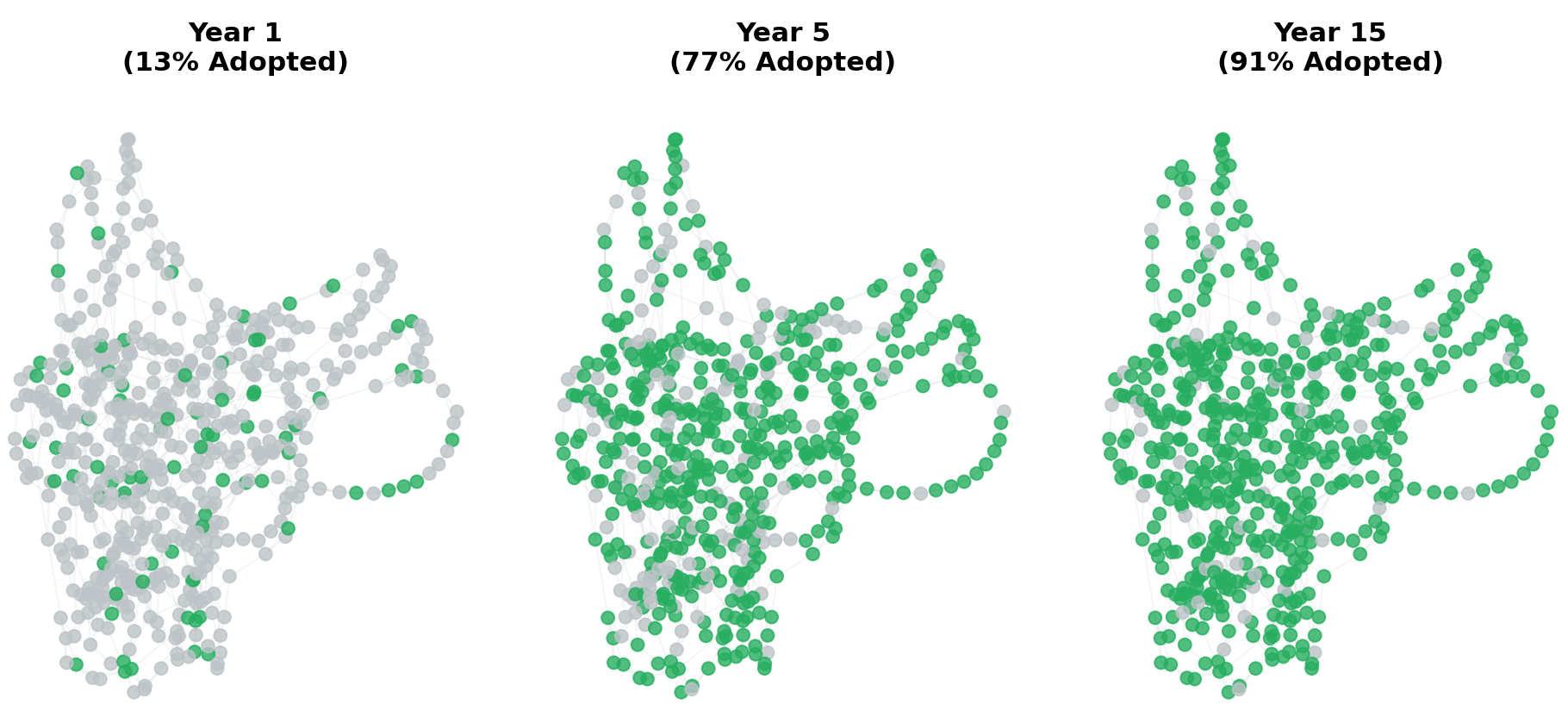}
    \caption{Longitudinal Network Snapshots of Spatial-Social Diffusion under a Subsidy Scenario (Years 1, 5, and 15).}
    \label{fig:Spatial_Social_Network_Diffusion}
\end{figure}

The longitudinal network snapshots (Fig.~\ref{fig:Spatial_Social_Network_Diffusion}) illustrate how technology diffusion evolves through the dairy sector’s social structure. The framework highlights that policy effects are path-dependent rather than purely immediate. Subsidies reduce initial costs while accelerating early diffusion, increasing visible adopters and strengthening peer validation. Adoption reaches 13\% in Year 1 and rises to 77\% by Year 5 as peer experience reduces perceived risk barriers. The model reaches approximately 91\% adoption by Year 15, with the remaining 9\% reflecting persistent resistance despite incentives. In contrast, carbon taxes mainly operate through ongoing cost penalties on conventional inputs; while reducing emissions, they may not generate the same self-reinforcing diffusion unless sufficiently strong to cross the social tipping point observed between Years 1 and 5.

\paragraph{Synthesis for Policy:}

Our results demonstrate that (i) emissions heterogeneity is accounted for by stocking imbalance and nutrient efficiency, (ii) social networks are significant in modeling realistic adoptions, and (iii) subsidy-based interventions create faster diffusion, greater cumulative abatement, and lower temporally derived climate risk than carbon taxes. Applying peer effects and minimising initial investment, near-total adoption can be achieved at very low marginal abatement cost to drive rapid steps towards national agricultural climate targets.

\section{Discussion}

The results demonstrate that heterogeneity in farm-level carbon intensity is structurally driven by differences in nutrient-use efficiency, land pressure, and managerial performance rather than farm scale alone. Empirical analysis shows that benchmark farms maintain high productivity with lower emissions, while the gap with laggard farms indicates significant potential for emission reductions.

The strong predictive accuracy of the model (Table~\ref{tab:validation}) and the substantial deterioration in performance when social networks are removed (Table~\ref{tab:network_test}) indicate that peer effects are a dominant mechanism within the modelled diffusion process. This finding supports the hypothesis that advisory strategies targeting influential farmers could accelerate sector-wide adoption.

Policy scenario analysis shows that financial incentives strongly influence diffusion timing and acceleration. The subsidy scenario achieves earlier peak adoption velocity (Fig.~\ref{fig:adoption_velocity}), increasing cumulative abatement over time (Fig.~\ref{fig:Cumulative_Greenhouse_Gas_Abatement}). Although both carbon tax and subsidy scenarios approach near-complete adoption (Fig.~\ref{fig:adoption_diffusion}), earlier adoption is critical for time-bound climate targets because emissions reductions accumulate over the simulation horizon. Distributional analysis (Fig.~\ref{fig:intensity_distribution_shift}) shows that policy-driven adoption reduces both mean carbon intensity and farm-level variation, indicating convergence towards best practice.

The marginal abatement cost of €0.80 per tCO$_2$e (Table~\ref{tab:abatement_costs}) is substantially below current carbon prices, indicating the potential cost-efficiency of fertiliser substitution within the model boundary. This estimate reflects fertiliser input cost differences only and excludes public subsidy expenditure, which requires separate welfare analysis. Monte Carlo simulations confirm the robustness of the subsidy scenario, with higher cumulative abatement ($p < 0.001$) and stable adoption outcomes under stochastic variation. Overall, the integration of empirical farm data with agent-based diffusion modelling demonstrates that adoption timing, social networks, and targeted incentives are important drivers of mitigation outcomes. Policies that reduce initial adoption barriers and leverage peer influence can accelerate emission reductions and support low-emission production systems in Irish dairy farming.

\paragraph{Path dependency and the heuristic value of non-linear scenario mapping:}An important implication of the framework is that policy effects may be path-dependent rather than purely contemporaneous. Subsidies reduce initial switching costs while accelerating early adoption, increasing visible adopters, strengthening peer validation, and shifting the system towards a higher adoption pathway. This transition can remain partially locked in because adoption is absorbing and perceived risks are reduced through peer experience. In contrast, carbon taxes mainly operate through continued price penalties on incumbent inputs, reducing emissions but generating weaker self-reinforcing diffusion unless sufficiently strong and persistent to cross the same social tipping point. Thus, the framework captures how policy timing and network effects shape durable and non-durable transition pathways.

\begin{table}[htp]
\centering
\scriptsize
\setlength{\tabcolsep}{4pt} 
\begin{tabular}{p{3.8cm} p{1.8cm} p{1.5cm} p{1.7cm} p{1cm} p{2cm} p{2cm}}
\hline
\textbf{Method / Study} & 
{\textbf{Accuracy}} & 
{\textbf{Social}\newline\textbf{Factors}} & 
{\textbf{Network}\newline\textbf{Dynamics}} & 
{\textbf{Multi-}\newline\textbf{Tech}} & 
{\textbf{Scalability}} & 
{\textbf{Validation}} \\
\hline
Our Study & 0.979 & Yes & Yes & Yes & High & Extensive \\
Alotibi2025\cite{Alotibi2025}  & 0.942 & Yes & Yes & Yes & High & Extensive \\
Williams2025\cite{Williams2025} & 0.876 & Partial & No & No & Medium & Limited \\
Campfens2025\cite{Campfens2025} & 0.834 & No & No & No & High & Extensive \\
Tech2024\cite{Tech2024} & 0.798 & Yes & No & No & Low & Case Study \\
Romero2024\cite{Romero2024} & 0.867 & Yes & Yes & No & Medium & Industry \\
Ates2024\cite{Ates2024} & 0.845 & Partial & No & No & Medium & Regional \\
Cardenas2025\cite{Cardenas2025} & 0.789 & No & No & Yes & Low & Theoretical \\
Collins2024\cite{Collins2024} & 0.856 & Partial & Yes & No & High & Synthetic \\
Egger2023\cite{egger2023} & - & Yes & No & No & Regional & Pattern-based \\ 
Javansalehi2024\cite{javansalehi2024} & 0.61 & Yes & No & Yes & Regional & Extensive \\ 
Farahbakhsh2025\cite{farahbakhsh2025} & - & Yes & Yes & No & Low & Sensitivity / Scenario \\ 
Tarruella2025\cite{tarruella2025} & - & Yes & Yes & No & Low & Sensitivity \\ 
\hline
\end{tabular}
\caption{Comprehensive comparison of agent-based and related adoption models in agriculture.}

\label{tab:method_comparison}
\end{table}

A comparison with existing agent-based and related adoption models (Table~\ref{tab:method_comparison}) shows that the present framework integrates empirical calibration, explicit social network structure, and multi-technology representation within a single modelling environment. While previous studies capture individual elements such as network diffusion, behavioural heterogeneity, or scenario analysis, few combine these components simultaneously. The higher predictive accuracy should therefore be attributed to the integration of empirical grounding and network-based diffusion rather than model complexity alone. This framework complements existing approaches by extending their theoretical and regional insights into agricultural technology adoption \cite{Alotibi2025}. By combining farm heterogeneity, social networks, and calibrated decision rules, the model provides a generalisable platform for analysing technology diffusion and emissions outcomes.  These results represent exploratory model-based insights rather than policy prescriptions, demonstrating the value of ABM for evaluating behavioural and structural drivers of mitigation adoption across technologies, regions, and incentive systems.

\subsection{Model Boundaries, Limitations, and Structural Transferability}

The framework provides insights into low-emission technology diffusion but is constrained by modelling assumptions and the institutional, socio-economic, and structural characteristics of the Irish dairy sector.

\paragraph{Internal Methodological Limitations:}
Agent decisions are driven by economic incentives and peer influence ($\omega$), excluding factors such as risk preferences, institutional trust, and contractual commitments. Fixed price differences between Calcium Ammonium Nitrate (CAN) and Protected Urea are assumed, while market variation and wider influences such as media, digital platforms, and advisory services are not captured.

\paragraph{Geographic and Institutional Transferability:}
The social network reflects the integrated and cooperative structure of Irish agriculture supported by the Teagasc extension framework. Regions with weaker advisory networks or fewer interactions would require adjustment of network density and contact frequency to avoid overestimating diffusion. Similarly, intensive or corporate dairy systems may require modified network structures due to more centralized decision-making.

\paragraph{Cross-Sectoral Applicability:}
Application to other agricultural sectors requires adjustments for farm structure, capital requirements, and labour dynamics. Drystock systems may require stronger incentives and slower diffusion, while tillage systems require consideration of technology costs, replacement cycles, and land tenure constraints.

\paragraph{Boundary Conditions and Implementation Costs:}
The reported private and social abatement costs (\texteuro0.80/tCO$_{2}$e and \texteuro4.54/tCO$_{2}$e) represent engineering baseline substitution costs rather than full macroeconomic estimates. They reflect direct CAN-Protected Urea price differences per unit of nitrogen and exclude broader implementation barriers affecting adoption.

To provide a realistic foundation for policy assessment, Table~\ref{tab:omitted_costs_framework} categorizes these unmodeled cost vectors and details how they systematically shift the true Marginal Abatement Cost Curve (MACC) upwards.

\begin{table}[htbp]
\centering
\caption{Systematic Overview of Unmodeled Behavioral and Transactional Implementation Costs.}
\label{tab:omitted_costs_framework}
\footnotesize
\setlength{\tabcolsep}{4pt}
\renewcommand{\arraystretch}{0.9}
\begin{tabular}{p{3cm} p{8cm} p{3.8cm}}
\hline
\textbf{Cost Category} & \textbf{Micro-Economic Mechanism} & \textbf{Expected MACC Impact} \\ \hline

\textbf{Transaction \& Search Costs} & Time and capital spent by farmers independently verifying agronomic safety, environmental compliance, and yield parity claims. & Marginal uniform shift upwards across all farm sizes. \\

\textbf{Advisory \& Extension Costs} & Public and private advisory infrastructure requirements (e.g., Teagasc multi-year discussion groups, soil sampling, nutrient tracking software). & Disproportionately increases social/public implementation costs. \\

\textbf{Learning \& Equipment Recalibration} & Technical labor time required to recalibrate spreading machinery disc speeds, preventing localized striping or grass-scorching. & Regressive impact; heavier burden on smaller, non-contractor farms. \\

\textbf{Behavioral Risk Premiums} & Subjective economic discount rates applied by operators to shield against perceived risks of yield penalties during unexpected weather shocks. & Acts as a steep initial adoption threshold barrier. \\

\textbf{Supply Chain Bottlenecks} & Inconsistencies in local cooperative retail stock inventory, forcing temporary substitution back to traditional CAN. & Introduces temporal adoption volatility. \\

\hline
\end{tabular}
\end{table}

Accounting for these frictional factors explains why real-world adoption often lags behind idealized economic simulations, a phenomenon known as the ``energy efficiency paradox'' or ``agronomic choice paradox.'' Although the transition appears cost-effective at purchase, infrastructure and behavioural risk mitigation required for widespread adoption mean that societal costs may exceed baseline estimates.

\section{Conclusion}
This study demonstrates the critical necessity of integrating farm-level heterogeneity, social network dynamics, and targeted policy instruments to understand the adoption of low-emission fertilizers in dairy systems. Using an agent-based model (ABM) calibrated with empirical data from 295 Irish farms, we demonstrate that adoption is fundamentally driven by peer-to-peer interactions and farm structural characteristics. Specifically, larger and more socially central farms act as 'opinion leaders,' accelerating the sector-wide diffusion of Protected Urea. Policy architecture significantly dictates these outcomes: subsidy-based interventions facilitate a faster transition and greater cumulative abatement than carbon taxes, with network effects further magnifying these gains. Economically, Protected Urea adoption is a highly efficient mitigation strategy; at a private cost of €0.80 per tCO2e, it is significantly more affordable than current carbon price benchmarks. While the model identifies a realistic saturation ceiling of 91\% by Year 15, the remaining 9\% 'laggard' margin highlights the persistent challenges of non-economic resistance. These findings offer a scenario-based perspective on diffusion rather than direct policy prescriptions. However, the results emphasize that targeting interventions at structurally influential, high-emission farms yields the greatest marginal mitigation impact. Ultimately, this integrated ABM framework provides an evaluative tool to analyze and design cost-effective, socially grounded decarbonization strategies for the agricultural sector.

Future development of this framework should explore: 1) multi-technology bundling to simulate adoption of complementary measures such as clover incorporation and low-emission slurry spreading; 2) dynamic behavioural profiling using heterogeneous agent archetypes to capture diverse decision-making beyond a uniform social contagion weight; 3) climate feedback integration by linking the ABM with biophysical models to assess adoption stability under extreme weather conditions; 4) cross-sectoral scalability by applying the framework to beef and tillage systems across regions with different network structures; and 5) real-time decision support through integration with current market prices and national statistics for dynamic emission-reduction forecasts.

\section*{Declarations}

\begin{itemize}

    \item \textbf{Funding Statement:} This research was supported by the Department of Agriculture, Food and the Marine (DAFM) 2023 Thematic Research Call (Project Reference: 2023RP956).

    \item \textbf{Authors’ Contributions:} S.J. conceived the study, designed and implemented the methodology, developed the Agent-Based Model, conducted the analyses, and wrote the manuscript. I.D. supervised the research, contributed to conceptual guidance, and reviewed the manuscript. K.S., J.M., and C.O. provided agricultural domain expertise. All authors reviewed and approved the final manuscript.

    \item \textbf{Acknowledgements:} The authors acknowledge financial support from the Department of Agriculture, Food and the Marine (DAFM) 2023 Thematic Research Call (Project Reference: 2023RP956).
    
    \item \textbf{Ethics Declaration:} Not applicable.

    \item \textbf{Consent to Participate:} Not applicable, as this study did not involve human participants.

    \item \textbf{Consent to Publish:} Not applicable, as the study does not contain individual-level personal data.

    \item \textbf{Informed Consent:} Not applicable.

    \item \textbf{Data Availability Statement:} Data will be made available on reasonable request.

    \item \textbf{Competing Interests:} The authors declare no competing interests.

    \item \textbf{Human Participants and/or Animals:} Not applicable.

    \item \textbf{Clinical Trial Registration:} Not applicable.

\end{itemize}

\renewcommand{\bibfont}{\footnotesize}
\setlength{\bibsep}{1pt}

\vfill

\end{document}